\documentclass[11pt,a4paper]{article}
\usepackage[hyperref]{emnlp2020}
\usepackage{times}
\usepackage{tabu}
\usepackage{latexsym}
\usepackage{amssymb}
\usepackage{graphicx}
\usepackage{booktabs}
\usepackage{multirow, tabularx}
\usepackage{caption}
\usepackage{subcaption}
\usepackage[export]{adjustbox}

\usepackage{microtype}

\aclfinalcopy 

\newcommand\inner[2]{\langle #1, #2 \rangle}

\title{Improve Transformer Models with Better Relative Position Embeddings}

\author{Zhiheng Huang \\
  AWS AI \\
  \texttt{zhiheng@amazon.com} \\\And
  Davis Liang \\
  AWS AI \\
  \texttt{liadavis@amazon.com} \\\AND
  Peng Xu \\
  AWS AI \\
  \texttt{pengx@amazon.com} \\\And
  Bing Xiang \\
  AWS AI \\
  \texttt{bxiang@amazon.com} \\  
  }

\date{}

\begin{document}
\renewcommand\baselinestretch{0.94}
\baselineskip=0.93\normalbaselineskip

\maketitle
\begin{abstract}
Transformer architectures rely on explicit position encodings in order to preserve a notion of word order.
In this paper, we argue that existing work does not fully utilize position information. For example, the initial proposal of a sinusoid embedding is fixed and not learnable. In this paper, we first review absolute position embeddings and existing methods for relative position embeddings. We then propose new techniques that encourage increased interaction between query, key and relative position embeddings in the self-attention mechanism. Our most promising approach is a generalization of the absolute position embedding, improving results on SQuAD1.1 compared to previous position embeddings approaches. In addition, we address the inductive property of whether a position embedding can be robust enough to handle long sequences.  We demonstrate empirically that our relative position embedding method is reasonably generalized and robust from the inductive perspective. Finally, we show that our proposed method can be adopted as a near drop-in replacement for improving the accuracy of large models with a small computational budget.
\end{abstract}

\section{Introduction}

The introduction of BERT \cite{devlin2018} has lead to new state-of-the-art results on various downstream tasks such as question answering and passage ranking. Variations of BERT, including RoBERTa \cite{liu2019-2}, XLNet \cite{yang2019}, ALBERT \cite{lan2019} and T5 \cite{raffel2019} have been proposed. At its core, BERT is non-recurrent and based on self-attention; in order to model the dependency between elements at different positions in the sequence, BERT relies on position embeddings. With BERT, the input embeddings are the sum of the token embeddings, segment embeddings, and position embeddings. The position embedding encodes the absolute positions from 1 to maximum sequence length (usually 512). That is, each position has a learnable embedding vector. The absolute position embedding is used to model how a token at one position attends to another token at a different position. 

Recent work suggested removing the next sentence prediction (NSP) loss with training conducted solely on individual chunks of text \cite{liu2019}. In this setup, the notion of absolute positions can be arbitrary depending on chunk start positions. Therefore, the association of a token to an absolute position is not well justified. Indeed, what really matters is the relative position or distance between two tokens $t_i$ and $t_j$, which is $j-i$. This phenomena has been realized and the relative position representation has been proposed in \citet{shaw2018,huang2018}, in the context of encoder decoder machine translation and music generation respectively.  \citet{shaw2018} has been modified in transformer-XL \cite{dai2019} and adopted in XLNet \cite{yang2019}. The relative position embedding in \cite{shaw2018} has been proven to be effective and thus it is adopted in \cite{raffel2019,song2020}. 

In this paper, we review the absolute position embedding from \citet{devlin2018} and the relative position
embeddings in \citet{shaw2018,dai2019}. Our contributions are as follows. 

\begin{enumerate}
\item We argue that the relative position is not fully utilized in the existing work. We propose a number of relative position embeddings in this paper in order to encourage increased interaction between the  key, query, and position embeddings. We show that our proposed embeddings can outperform the widely used relative position embedding \cite{shaw2018} on SQuAD1.1.  
\item We discuss the \textit{inductive property}: can BERT, trained on short sequences, generalize to longer sequences from the perspective of position embeddings? We conduct ablation studies to show how the clipping value $k$ (used to limit the relative distance) affects the model accuracy. We demonstrate empirically that our relative embedding method is robust with respect to this \textit{inductive property}. 
\item We show that our novel position embedding technique can improve BERT-large performance with only a few epochs of fine-tuning. Acquiring large gains with a small computation budget.
\end{enumerate}

\section{Related Work}

Previously, \citet{aswani2017} introduced a position embeddings with dimensions matching the token embeddings (so that they can be summed). Specifically, they choose the sine and cosine functions at different frequencies:
\begin{eqnarray}
PE_{(pos, 2i)} = sin(pos/10000^{2i/d_{model}}) \\
PE_{(pos, 2i+1)} = cos(pos/10000^{2i/d_{model}})
\end{eqnarray}
where $pos$ is the position and $i$ is the embedding dimension. That is, each dimension of the position encoding corresponds to a sinusoid. The authors hypothesized that it would allow the model to easily learn to attend via relative positions, since for any fixed offset $k$, $PE_{pos+k}$ can be represented as a linear function of $PE_{pos}$. They also experimented with learned position embeddings \cite{gehring2017} and found that the two versions produced nearly identical results. BERT \cite{devlin2018} uses a learnable position embedding.

Previous work \cite{parikh2016} has introduced attention weights based on relative distance prior to BERT \cite{devlin2018}. More recently, \citet{shaw2018} demonstrated the importance of relative position representations. They presented an efficient way of incorporating relative position representations into the transformer self-attention layer. They achieved significant improvements in translation quality on two machine translation tasks. \citet{huang2018} has proposed a similar idea to incorporate the relative distance explicitly but in the music generation domain.  Transformer-XL \cite{dai2019} has modified \cite{shaw2018} to have the following two differences: 1) to introduce additional bias terms for queries; and 2) to re-introduce the use of a sinusoid formulation, in the hope that a model trained on a memory of a certain length can automatically generalize to a memory several times longer during evaluation\footnote{This was not rigorously verified in experiments.}. The proposed relative position embedding has been used in transformer-XL \cite{dai2019} and XLNet \cite{yang2019}. The relative position embedding by \citet{shaw2018} is proven to be effective and it is validated in BERT variants model training \cite{raffel2019,song2020}. 

In addition to the above work, \citet{chorowski2015} proposed a novel method of adding location-awareness to the attention mechanism in the sequence to sequence framework for automatic speech recognition (ASR). Their work is related to this paper as both attempt to integrate a location information into the self-attention mechanism.

\section{Position Embeddings}
In this section, we review the absolute position embedding used in the original BERT paper and the relative position embedding proposed in \cite{shaw2018,dai2019}. We then propose a number of relative position embeddings, from simpler ones to more complex ones. We analyze the complexity of each embedding method. 

\subsection{Self-Attention review}
\begin{figure}[!hbt]
    \centering
    \includegraphics[width=1.1\columnwidth]{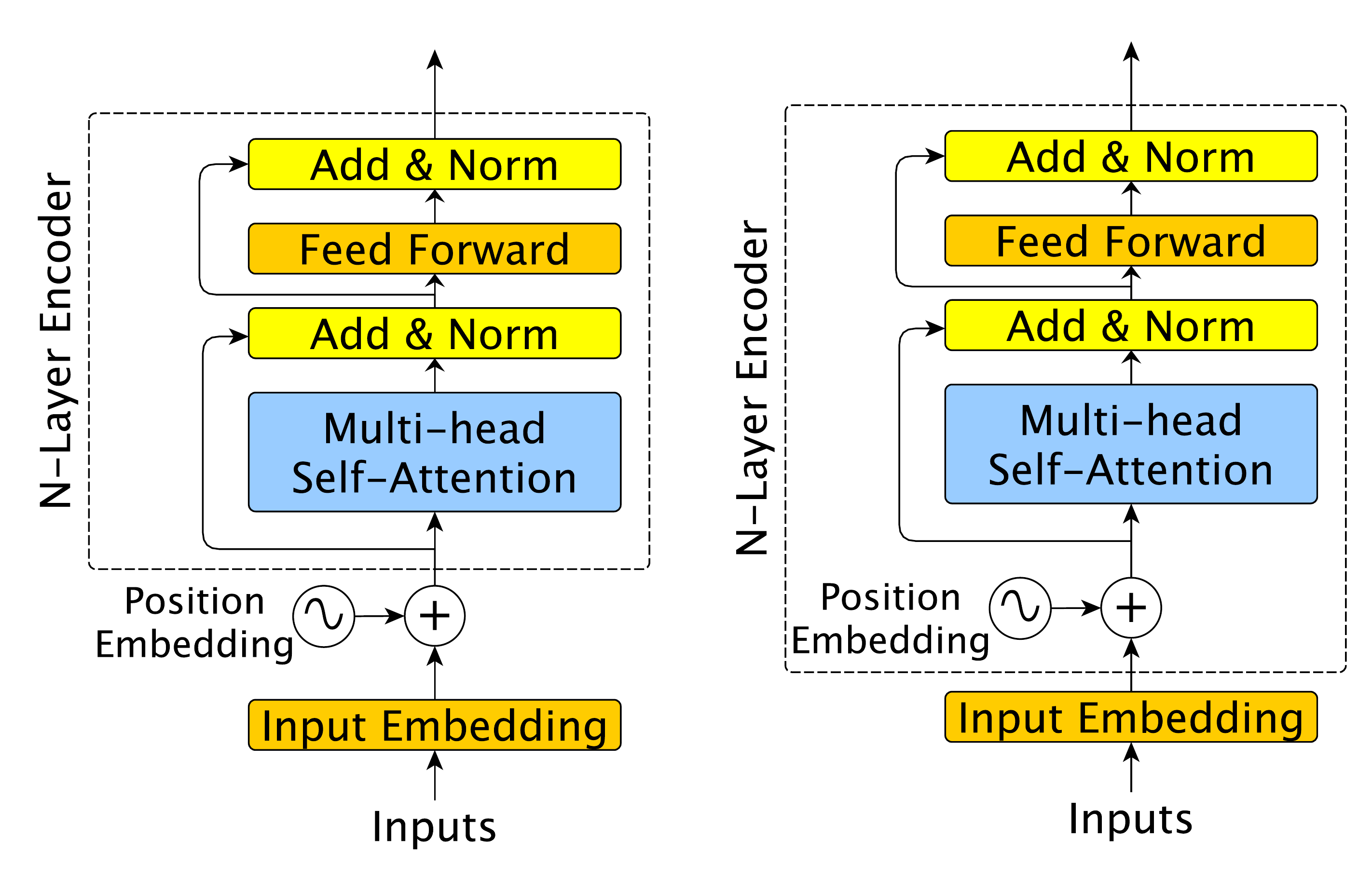}
    \caption{Transformer architectures with the original absolute position embedding (left) and all other variations of relative position embeddings (right).}
    \label{fig:trans}
\end{figure}

The BERT model consists of a transformer encoder \cite{aswani2017} as shown in Figure \ref{fig:trans}. 

The original transformer architecture uses multiple stacked self-attention layers and point-wise fully connected layers for both the encoder and decoder. Each self-attention sublayer consists of $h$ attention heads. The result from each head are concatenated to form the sublayer's output. Each attention head operates on an input sequence, $x=(x_1, \ldots, x_n)$ of $n$ elements (maximum number of tokens allowed in model training, $n$ is usually 512 in default) where $x_i \in \mathbb{R}^{d_x}$, and computes a new sequence $z = (z_1, \ldots, z_n)$ of the same length where $z_i \in \mathbb{R}^{d_z}$. Each output element, $z_i$, is computed as weighted sum of linearly transformed input elements:
\begin{equation}
z_i = \sum_{j=1}^{n} \alpha_{ij}(x_jW^V),
\end{equation}
where $\alpha_{ij}$ is the weight which is computed by applying a softmax function:
\begin{equation}
\alpha_{ij} = \frac{\exp e_{ij}}{\sum_{k=1}^n \exp e_{ik}},
\end{equation}
where $e_{ij}$ is the attention weight from position $j$ to $i$, a scaled dotted product following a linear transformation:
\begin{equation}
\label{eqn:e}
e_{ij} = \frac{(x_i W^Q)(x_j W^K)^T}{\sqrt{d_z}}.
\end{equation}
The scaling factor $\sqrt{d_z}$ is necessary to make the training stable. The dot product is chosen due to its simplicity and computational efficiency. Linear transformation of the inputs add sufficient expressive power. $W_Q$, $W_K$, $W_V \in \mathbb{R}^{d_x \times d_z}$ are parameter matrices. These parameter matrices are unique per layer and attention head.

\subsection{Absolute position embedding in BERT}
\begin{figure}[!hbt]
    \centering
    \includegraphics[width=0.8\columnwidth]{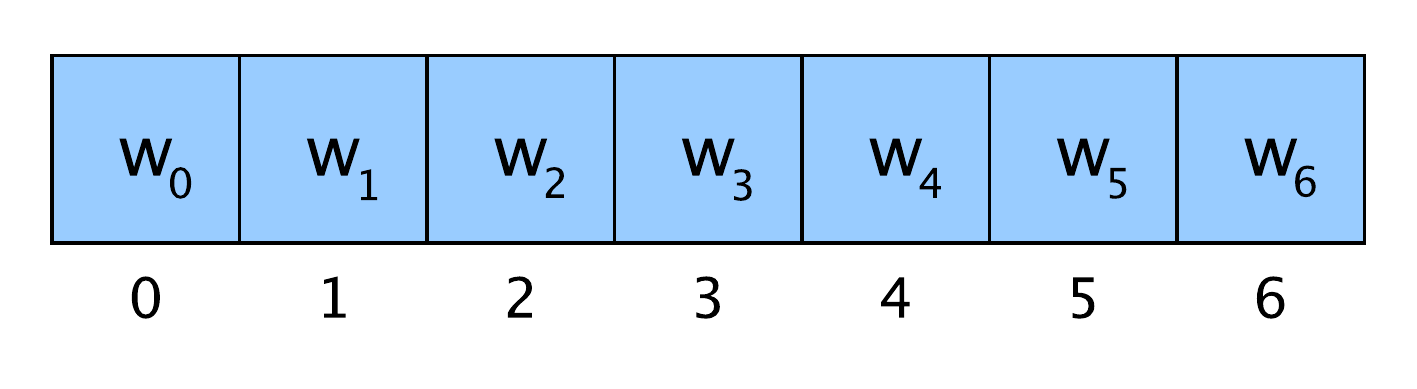}
    \caption{Absolute position embedding $p_{i}$.}
    \label{fig:absolute_weights}
\end{figure}

In the self-attention scheme, the absolute position embedding is as follows.
\begin{equation}
\label{eq:ab_pos}
x_i = t_i + s_i + w_i,
\end{equation}
where $x_i$, $i \in \{0, \ldots, n-1\}$ is the input embedding to the first transformer layer, $t_i$, $s_i$ and $w_i \in \mathbb{R}^{d_x}$ are the token embeddings, segment embeddings and absolute position embeddings respectively.  Segment embedding indicates if a token is sentence $A$ or sentence $B$, which was originally introduced in BERT \cite{devlin2018} to compute the next sentence prediction (NSP) loss. Later work \cite{yang2019,liu2019,raffel2019} suggested that the NSP loss does not help improve accuracy. We therefore drop the segment embedding in this paper. Token embeddings $t_i$ and absolute position embeddings $w_i$, are learnable parameters trained to maximize the log-likelihood of the MLM task. Figure \ref{fig:absolute_weights} depicts the absolute position embedding graphically, which is used in the first layer in Figure \ref{fig:trans} left. The maximum length of a sequence $n$ is required to be determined before the training. Although it lacks the inductive property, this approach is found to be effective for many NLP tasks, due to the fact that the maximum sequence length is enforced at inference anyway in most cases.

\subsection{Shaw's relative position embedding}
The work of \cite{shaw2018} proposed the edge representations, $a_{ij} \in \mathbb{R}^{d_z}$, which is used to model how much token $t_i$ attends to token $t_j$. The equation (\ref{eqn:e}) can be revised as follows to consider the distance between token $i$ and $j$ in computing their attention.
\begin{equation}
e_{ij} = \frac{(x_i W^Q)(x_j W^K + a_{ij})^T}{\sqrt{d_z}}.
\end{equation}

\begin{figure}[!hbt]
    \centering
    \includegraphics[width=0.7\columnwidth]{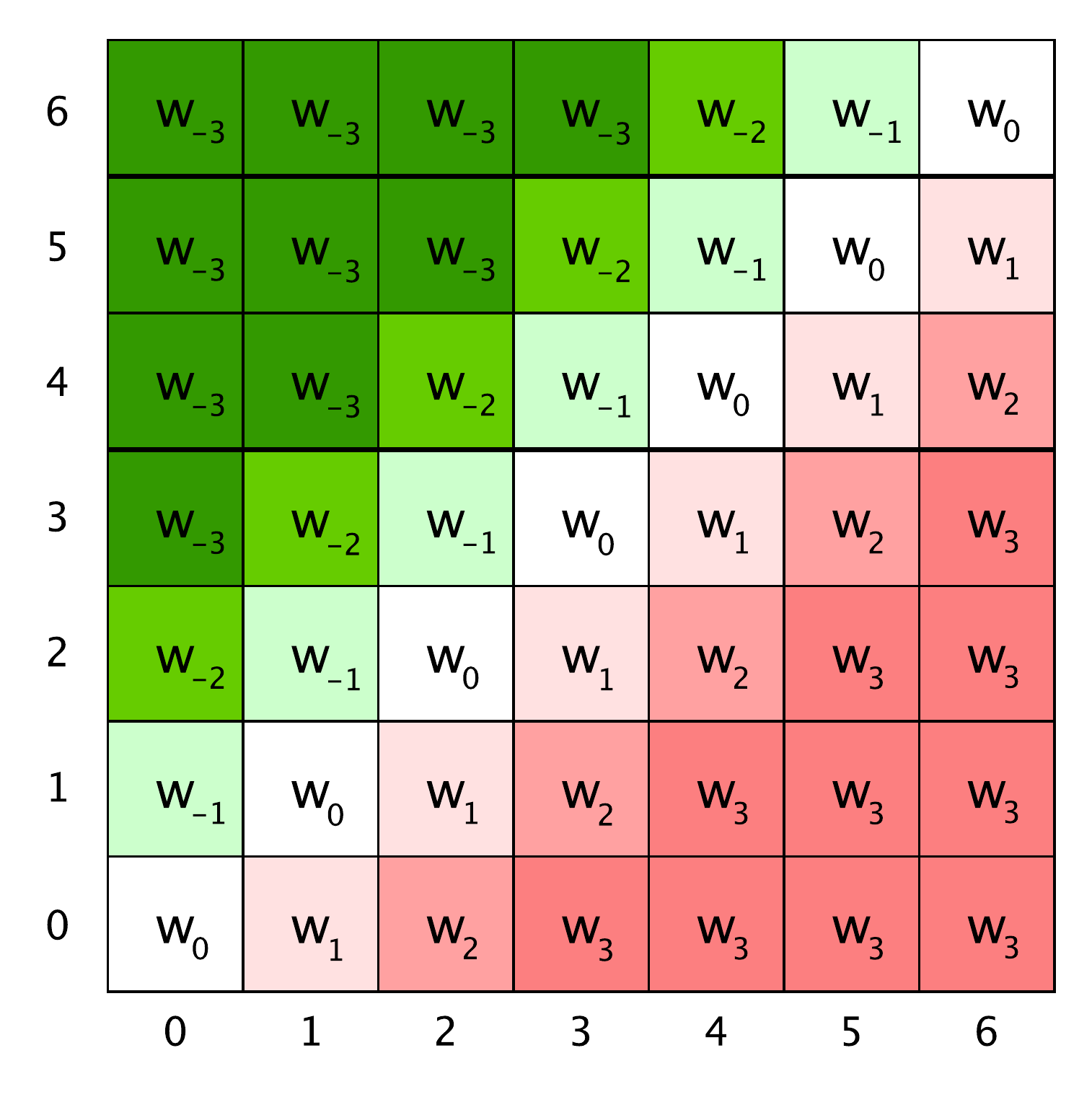}
    \caption{Relative position weights $a_{ij}$.}
    \label{fig:relative_weights}
\end{figure}

They also introduced clipped value $k$ which is the maximum relative position distance allowed. The authors hypothesized that the precise relative position information is not useful beyond a certain distance. Therefore, there are $2k+1$ unique edge labels $w = (w_{-k}, \ldots, w_k)$ defined as the following.
\begin{eqnarray}
a_{ij} &=& w_{\mathrm{clip}(j-i,k)} \label{eqn:aij}\\
\mathrm{clip}(x,k) &=& \max (-k, \min(k,x))  \label{eqn:k}
\end{eqnarray}
Figure \ref{fig:relative_weights} shows the edge representations $a_{ij}$ graphically, with $k=3$. 

\subsection{XLNet's relative position embedding}
Transformer-XL \cite{dai2019} and XLNet \cite{yang2019} also utilize the relative position embedding, with the equation (\ref{eqn:e}) being revised as follows
\begin{equation}
\label{eqn:e_xl}
\small
e_{ij} = \frac{(x_i W^Q + u)(x_j W^K )^T  +  (x_i W^Q + v) (R_{ij}W^R)^T}{\sqrt{d_z}},
\end{equation}
where $W^R$ is a learnable parameter matrix and $R_{ij}$ is the sinusoid encoding vector between location $i$ and $j$.  $R$ is a sinusoid encoding matrix \cite{aswani2017} without learnable parameters, which essentially reflects the prior that only the relative distance matters for where to attend. $u \in \mathbb{R}^{d_z}$ and $v \in \mathbb{R}^{d_z}$  are trainable parameters to represent the query bias for content-based (the first term in numerator) and location-based (the second term in numerator) attentions respectively. The relative position embedding defined in equation (\ref{eqn:e_xl}) is similar to the work of \citet{shaw2018} but with two differences: 1) it introduces additional bias terms for queries; and 2) it uses the sinusoid formulation proposed in the original transformer paper \cite{aswani2017}.

We implemented this but found that the bias terms led to training instability. After removing the bias terms, keeping only the sinusoids, we found that the accuracy is slightly worse than Shaw's method \cite{shaw2018}. We skip the comparison to XLNet's relative embedding while focusing on the comparison to the Shaw's method, which has been widely used in the variants of BERTs due to its simplicity \cite{raffel2019,song2020}.

\subsection{Proposed position embeddings}
In this section, we propose four variants of relative position embedding to encourage increased interactions between key, query, and position embedding in the self-attention mechanism. The design choices include whether relative positions are signed and whether they are scalars or vectors. 

\subsubsection{Relative position embedding method 1}
This method only considers the absolute distance of token $i$ and $j$. That is, it does not distinguish the sign of the distance $j-i$. The distance embedding can be written as follows.
 \begin{equation}
a_{ij} = w_{|j-i|}, 
\end{equation}
where $w$ is scalar used to represent how token $i$ attends to $j$ with absolute distance $|j-i|$. We do not apply the clipping value $k$ in this method. The learnable parameters are $w = (w_{0}, \ldots, w_{n-1})$, where $n$ is maximum sequence length. The equation (\ref{eqn:e}) can be revised as follows to consider the distance between token $i$ and $j$ in computing their attention. As $a_{ij}$ is a scalar, we use the multiplicative interaction between key, query and relative embedding, which is different from the additive interaction in Shaw's method.
\begin{equation}
\label{eqn:e_2}
e_{ij} = \frac{(x_i W^Q)(x_j W^K )^Ta_{ij}}{\sqrt{d_z}}.
\end{equation}

\subsubsection{Relative position embedding method 2}
As with method 1, this method uses scalars to represent relative position embeddings. However, it now distinguishes the sign of the distance $j-i$. That is, it assumes that the future token has different attention weights from the previous one in attending to a token in the middle, despite that the absolute distance is the same. The distance embedding can thus be written as follows.
 \begin{equation}
a_{ij} = w_{j-i}, 
\end{equation}
where $w$ is scalars used to represent how token $i$ attends to $j$. The learnable parameters are $w = (w_{1-n}, \ldots, w_{0}, \ldots, w_{n-1})$, where $n$ is maximum sequence length. Similar to method 1, the equation (\ref{eqn:e_2}) is used compute the attention scores.

\subsubsection{Relative position embedding method 3}
Method 3 replaces the scalar relative position embeddings with vector embeddings. The distance embedding can thus be written as follows.
 \begin{equation}
a_{ij} = \mathbf{w}_{j-i}, 
\end{equation}
where $\mathbf{w} \in \mathbb{R}^{d_z}$  represents the embedding on how token $i$ attends to $j$. The learnable parameters are $\mathbf{w} = (\mathbf{w}_{1-n}, \ldots, \mathbf{w}_{0}, \ldots, \mathbf{w}_{n-1})$, where $n$ is maximum sequence length. The equation (\ref{eqn:e}) can be revised as follows.
 \begin{equation}
 \label{eqn:e_3}
e_{ij} = \frac{\mathrm{sum\_prod}(x_i W^Q, x_j W^K, a_{ij})}{\sqrt{d_z}}.
\end{equation}
Note that the numerator is the sum over element-wise product of three vectors in dimension $\mathbb{R}^{d_z}$: query vector, key vector and relative position embedding vector. This is a natural extension from multiplication of scalars in method 2. The key difference is the introduction of the multiplicative interaction between key, query, and the relative position \textit{vector}, which was missing in all previous methods (including absolute position embeddings and \citet{shaw2018} and XLNet's relative position embeddings). For example,  in \citet{shaw2018}, equation (\ref{eqn:e}), the attention score has two factors. The first models the interaction between key and query,  $(x_i W^Q)(x_j W^K)^T$, and the second models the interaction between query and relative position embedding, $(x_i W^Q)(a_{ij})^T$. We hypothesize that the explicitly modeling of the interaction between query, key and relative position embedding would have more expressive power. In this method, the relative position embedding serves as a gate to filter out the dot product of query and key. This gate would prevent a query from attending to a similar key (content-wise) heavily if the query and key positions are far away from each other. 

\subsubsection{Relative position embedding method 4}
We identified that all previous relative position embeddings do not model the interaction of query, key and relative position embeddings simultaneously. As a backoff from method 3 and also an extension to Shaw's method, method 4 consists of modeling the dot product of all possible pairs of query, key, and relative position embeddings. As with method 3, the learnable parameters are $\mathbf{w} = (\mathbf{w}_{1-n}, \ldots, \mathbf{w}_{0}, \ldots, \mathbf{w}_{n-1})$. The equation (\ref{eqn:e}) can be revised as follows.
 \begin{equation}
 \label{eqn:e_4}
 \small
e_{ij} = \frac{(x_i W^Q) \cdot (x_j W^K )  +  (x_i W^Q) \cdot a_{ij} + (x_j W^K ) \cdot a_{ij} }{\sqrt{d_z}}.
\end{equation}
Three factors in the numerator model the interaction of query and key, query and relative position embedding, and key and relative position embedding, respectively. The interaction of query and key is the conventional content attention, while the remaining two are for relative position discount of query and key respectively. Shaw's method (see equation \ref{eqn:e}) only contains the first two factors. We note that the embeddings are shared in factor 2 and 3, the formulation in (\ref{eqn:e_4}) empowers a more reliable estimation of relative embeddings compared to Shaw's method, as we will see in the experiments. Method 4 can be re-written as, 
\begin{equation}
e_{ij} = \frac{(x_i W^Q + a_{ij})(x_j W^K + a_{ij})^T - \inner{a_{ij}}{a_{ij}}}{\sqrt{d_z}}.
\end{equation}
The first term is a generalized case to absolute position embeddings (see equation (\ref{eq:ab_pos})), in which each absolute position embedding vector is added to the word embedding. Precisely, the assignment of $a_{ij} = a_{i}$ and $a_{ij} = a_{j}$ for the two entries of $a_{ij}$ in the first factor and the drop of the bias term $\inner{a_{ij}}{a_{ij}}$ make absolute position embeddings a specific case of method 4.

\subsection{Complexity Analysis}
We analyze the storage complexity of various position embedding methods in this section. For a transformer model with $m$ layers, $h$ attention heads per layer, and maximum sequence length of $n$, table \ref{tab:complexity} lists the parameter size for various position embeddings and the runtime storage complexity.  In order to have sufficient expressive power, we allow different embedding parameters at different layers for all methods  (see Figure \ref{fig:trans} right) except absolute position embedding\footnote{To be compatible to the original BERT implementation.}. For example, Shaw's method introduces relative position embedding parameters with size of $mh(2n-1)d$. The parameters are used multiple times in equation (\ref{eqn:aij}) (also see Figure \ref{fig:relative_weights}), leading to runtime storage complexity of $\mathcal{O}(mhn^2d)$. 

\begin{table}[!hbpt]
    \small
    \centering
    \begin{tabular}{lcc}
    Method & Parameter size & Complexity \\ \hline
    Absolute & $nd$ & $\mathcal{O}(nd)$ \\
    Shaw & $mh(2n-1)d$ & $\mathcal{O}(mhn^2d)$ \\
    method 1 & $mhn$ & $\mathcal{O}(mhn^2)$ \\
    method 2 & $mh(2n-1)$ & $\mathcal{O}(mhn^2)$ \\
    method 3 & $mh(2n-1)d$ & $\mathcal{O}(mhn^2d)$ \\
    method 4 & $mh(2n-1)d$ & $\mathcal{O}(mhn^2d)$ \\
    \hline
    \end{tabular}
    \caption{Parameter sizes and runtime storage complexities of various position embedding methods. }
\label{tab:complexity}
\end{table}
All position embedding methods introduce a small number of additional parameters to the BERT model. Precisely, Shaw, method 3 and 4 introduce  $mh(2n-1)d$, $12*12*(2*512-1)=147K$ parameters at maximum, which is negligible when compare to the number of parameters in BERT (108M parameters). For simple methods 1 and 2, they introduce even fewer parameters. We point out a caveat on method 3: despite the fact that it introduces the same number of parameters as with method 4, it requires a significantly higher memory footprint during training. This may be due to the inefficient GPU implementation of sum over element-wise product of vectors in Equation (\ref{eqn:e_3}) compared to matrix multiplication. As a result, we can only fit 2 sequences in each GPU for method 3, as opposed to 20 sequences per GPU for all other methods. In terms of training and inference speed, Shaw's method and proposed methods 1, 2 and 4 are all similar to the absolute position embedding baseline.

\section{Experiments}

We leverage the same data used to pre-train BERT: BooksCorpus (800M words) \cite{zhu2015} and English Wikipedia (2.5B words) \cite{wiki2004,devlin2018}. Following the setup from RoBERTa  \cite{liu2019}, we leave out the next sentence prediction loss and only use one segment instead of the two segments proposed in BERT \cite{devlin2018} during model training. We set the maximum input length to 512. Similar to BERT, we use a vocabulary size of 30k with wordpiece tokenization. 

We generate the masked input from MLM targets using \textit{whole word masking}. The model updates use a batch size of $160$ and Adam optimizer with learning rate starting at 1e-4. Our maximum batch size is 160 on an Nvidia V100 instance (with 8GPUs). 

Following previous work \cite{devlin2018,yang2019,liu2019,lan2019}, we evaluate on the General Language Understanding Evaluation (GLUE) benchmark \cite{wang2018} and the Stanford Question Answering Dataset (SQuAD1.1) \cite{rajpurkar2016}. 

\subsection{Models evaluation on SQuAD dataset} \label{sec:exp_squad}
We run the pre-training experiments for different position embedding methods on base settings only. We omit the BERT-large experiments as they are computationally intensive. After pre-training, we fine-tune on SQuAD1.1. Table \ref{tab:squad} shows the results of SQuAD for absolute position embedding, Shaw's relative position embedding, and the four relative position embeddings proposed in this paper.
\begin{table}[!hbpt]
\begin{minipage}{1.0\linewidth}
	\centering
	\footnotesize
\begin{tabu}{lccc}
  \toprule
  \textbf{Model} & \textbf{EM} & \textbf{F1} \\ 
  \midrule
  BERT \cite{devlin2018} &  80.8 & 88.5 & \\
  Absolute & 81.58 & 88.59 \\
  \cite{shaw2018} & 82.38 & 89.37 \\
  \midrule
  Method 1 & 80.82 & 87.96 \\
  Method 2 & 81.44 & 88.86 \\
  Method 3 & \textbf{83.71} & 90.50 \\ 
  Method 4 & 83.63 & \textbf{90.53} \\ 
  \bottomrule
\end{tabu}
\end{minipage}
\caption{SQuAD1.1 development results for various position embeddings on BERT-base.}
\label{tab:squad}
\end{table} 

We reproduced compatible BERT baselines \cite{devlin2018} (F1 score of 88.5) with absolute position embedding (F1 score of 88.59). We show that Shaw's relative position embedding leads to a higher accuracy (F1 score of 89.37) when compared to the BERT default setting (absolute position embedding). Our proposed simple relative position embeddings method 1 results in F1 scores of 87.96, which is worse than the baseline of absolute position embedding. When we consider the relative distance sign (method 2), we obtain an improved F1 score of 88.86, which is similar to the BERT absolute position embedding baseline. This shows the effectiveness of multiplicative interaction between query, key and relative embedding directly, despite that the relative embeddings are simple scalars. The method 3, which has vector representations for relative position embeddings and also models the interaction between query, key and relative position embedding directly, leads to a higher F1 score of 90.50. Finally, the method 4, which is backoff of method 3 (or extension of Shaw's method), leads to a similar F1 score of 90.53. Method 4 is the most promising method among four proposed methods due to its high accuracy and computation efficiency.

\subsection{Model evaluation on GLUE datasets}
Following \citet{devlin2018}, we use a batch size of 32 and 3-epoch fine-tuning over the data for GLUE tasks. For each task, we report the accuracy on development dataset with learning rate 3e-5. Table \ref{tab:glue} shows the results of GLUE datasets for absolute position embedding, Shaw's relative position embedding and the four proposed methods in this paper. 
\begin{table*}[!hbpt]
    \centering
    \small
    \begin{tabular}{lcccc}
    \toprule
    \textbf{Model} & \textbf{MNLI-(m/mm)} & \textbf{QQP} & \textbf{SST-2} & \textbf{MRPC}   \\ 
    \midrule
    BERT \cite{devlin2018} & 84.6/83.4 & 71.2 &  \textbf{93.5}  &  \textbf{88.9} \\
    Absolute & 83.57/83.65 & 87.64 &  90.48  &  88.40 \\
    \cite{shaw2018} & 84.10/84.07 & \textbf{87.77} &  90.94  &  88.68 \\
    \midrule
    Method 1 & 83.84/84.06 & 87.52 &  91.97  &  88.65 \\
    Method 2 & 83.68/83.78 & 87.50 &  91.05  &  87.34 \\
    Method 3 & \textbf{84.81}/\textbf{84.68} & 87.11 &  91.39  &  82.86 \\
    Method 4 & 84.45/84.51 & 87.41 &  91.74  &  88.88 \\
    \bottomrule
    \end{tabular}
    \caption{GLUE development results for different position embeddings on BERT base setting. }
\label{tab:glue}
\end{table*}

Following the settings from BERT \cite{devlin2018}, F1 scores are reported for QQP and MRPC, and accuracy scores are reported for MNLI and SST-2. There is no significant accuracy difference between the absolute, Shaw and proposed methods, except that the proposed method 3 leads to significant lower F1 score (82.86) on MRPC dataset. While various position embeddings lead to different results on complex question answering datasets like SQuAD, they are not as sensitive to GLUE tasks. Our hypothesis is that SQuAD requires hidden activations at all token positions, so the relative position embedding plays a key role in modeling the interactions of tokens at different positions. The GLUE datasets, on the other hand, use the first token [CLS] only and thus the relative embeddings have limited impact. We do not know the exact reason for the low accuracy of method 3 on MRPC dataset. One hypothesis is that the interaction between query, key and position embedding introduced in method 3 is unstable on this dataset. 

\subsection{Models with various $k$}
We usually limit the maximum training sequence length to 512 in BERT training in consideration of the memory footprint. It remains relatively unexplored for the \textit{inductive} property: can a BERT model trained on short sentences be generalized to handle longer sentences? This property is not thoroughly explored, partially because a maximum sequence length would be applied during inference anyway for practical considerations and thus there is no consistency between training and testing. Nevertheless, to fully address the question, one can train BERT models with different settings of maximum sequence lengths and test on longer sequences. The inductive property is related to the position embedding methods. One can try different position embedding methods and test how they affect the inductive property. For example, if we set $m$ and $n$, $m < n$, as the maximum sequence lengths for training and test respectively. The fixed sinusoid, Shaw's, and our proposed methods can be directly employed while the absolute position method cannot as the position embeddings for position $[m+1, n]$ are not learned in training but are required during inference. The relative position embeddings are better choices as they are not subject to a maximum position value and learnable. In this section, we vary the clipping distance $k$ (equation \ref{eqn:k}) of the maximum relative distance to see how it affects the model accuracy. A small value of $k$ explicitly models two tokens within this distance. Any pairs of tokens greater than this would be treated as if they are $k$ positions away. 

Table \ref{tab:squad_k} shows the EM and F1 score of method 4 on SQuAD dev dataset as a function of $k$. 
\begin{table}[!hbpt]
\centering
\small
    \begin{tabular}{lccccc}
    \toprule
   & \textbf{Pre-train} & \textbf{Fine-tune} &  &  \\ 
  \textbf{k}  & \textbf{MaxSeqLen} & \textbf{MaxSeqLen} & \textbf{EM} & \textbf{F1} \\
  \midrule
 2 & 512 & 512 & 82.17 & 89.19 \\ 
 4 & 512 & 512 & 82.44 & 89.37 \\ 
 8 & 512 & 512 & 82.64 & 89.59 \\ 
 16 & 512 & 512 & 83.42 & 90.18 \\ 
  32 & 512 & 512 & 83.58 & 90.30 \\ 
  64 & 512 & 512 & 83.40 & 90.21 \\ 
 128 & 512 & 512 & 83.20 & 90.03 \\ 
  256 & 512 & 512 & 83.59 & 90.54 \\ 
  512 & 512 & 512 & \textbf{83.63} & \textbf{90.53} \\ 
  \midrule
  256 & 512 & 576 & 83.80 & \textbf{90.71} \\ 
  256 & 512 & 640 & \textbf{83.97} & 90.68 \\ 
  256 & 512 & 704 & 83.44 & 90.32 \\
  \bottomrule
  \end{tabular}
    \caption{SQuAD development results for different $k$ on BERT base setting.}
\label{tab:squad_k}
\end{table} 
The SQuAD dev data consists of 10570 question answer pairs. The average lengths of training and development sequences (questions and documents) are 130 and 133 respectively. We observe that the accuracy on SQuAD dev remains similar with $k \ge 32$\footnote{Note that in \citet{shaw2018}, they found that the BLEU scores remains the same when $k \ge 2$ in encoder decoder architecture.}. This suggest that the relative position embedding for token pairs are greater than $k=32$ can only provide marginal  information in BERT model training even the training sequences consist of 130 tokens in average. This observation ensures that method 4 is robust and generalized from the inductive perspective.

The absolute position embedding used in BERT does not permit downstream fine-tuning tasks training on sequences which have more tokens than the maximum sequence length (512). This, however, is not an issue for relative position embedding methods proposed in this paper. We hypothesize that this flexibility may offer further accuracy boost on downstream tasks. We fine-tune the model, which was pre-trained with $k=256$ and pre-train maximum sequence length of 512, on SQuAD training data but allowing increased maximum sequence lengths (576, 640 and 704 respectively)\footnote{Around 3\% of sequences have more tokens than maximum sequence length of 512.}. The bottom of Table \ref{tab:squad_k} confirms our hypothesis. For example, setting fine-tuning maximum sequence length to 576 results in the highest F1 score ($90.71\%$). We note that the gain mainly comes from the small percentage of SQuAD training and test data which have more tokens than 512. We hypothesize that, for a dataset which have a large percentage of sequences with 512 or more tokens, the gain can be more significant.

\begin{table*}[!hbpt]
\centering
\small
    \begin{tabular}{lccccc}
  \toprule
  & \textbf{Pre-train} & \textbf{Fine-tune} & & \\
  \textbf{Model} & \textbf{MaxSeqLen} & \textbf{MaxSeqLen} & \textbf{EM} & \textbf{F1} \\ \hline
  \midrule
  BERT \cite{devlin2018} & 512 & 512 & 84.1 & 90.9 & \\
  Pre-trained & 512 & 512 & 86.91 & 93.15 \\
  Method 4 & 512 & 512 & \textbf{87.51} & \textbf{93.55} \\ \hline
  Method 4 & 512 & 576 & 87.40 & 93.48 \\
  Method 4 & 512 & 640 & \textbf{87.57} & 93.49 \\
  Method 4 & 512 & 704 & 87.16 & 93.47 \\
  \bottomrule
  \end{tabular}
    \caption{SQuAD development results for various BERT models for large setting.}
\label{tab:squad_large}
\end{table*}

\subsection{Relative position embeddings for large BERT models}
Training BERT large models is computational expensive. To minimize the training cost, we test method 4 on a pre-trained BERT large model. In particular, we load a pre-trained BERT large model, \textit{bert-large-uncased-whole-word-masking}, from pytorch transformer\footnote{https://github.com/huggingface/transformers} as the initial model and fine-tune the existing parameters and the new relative position embedding parameters for 3 epochs, staring with a small learning rate of $5e-5$. We do not do the clipping of relative distance (thus $k=512$). Table \ref{tab:squad_large} shows that, with the near drop-in replacement of absolute position embedding with relative position embedding, method 4 boosts the F1 score of 93.15 to 93.55, with negligible increased number of parameters and inference latency. We hypothesize the same fine-tuning can be applied to different BERT variants (e.g., RoBERTa) to boost new state-of-the-art accuracy. In addition, we allow larger maximum sequence lengths (576, 640 and 704) in SQuAD fine-tuning task but do not observe additional gain in this case.

\subsection{Relative Position Visualization}
We attempt to visualize relative position embeddings in this section. We select method 4 for visualization as it is both the most efficient and most accurate amongst our proposed methods. 

\begin{figure}[!hbt]
    \centering
    \includegraphics[scale=0.3]{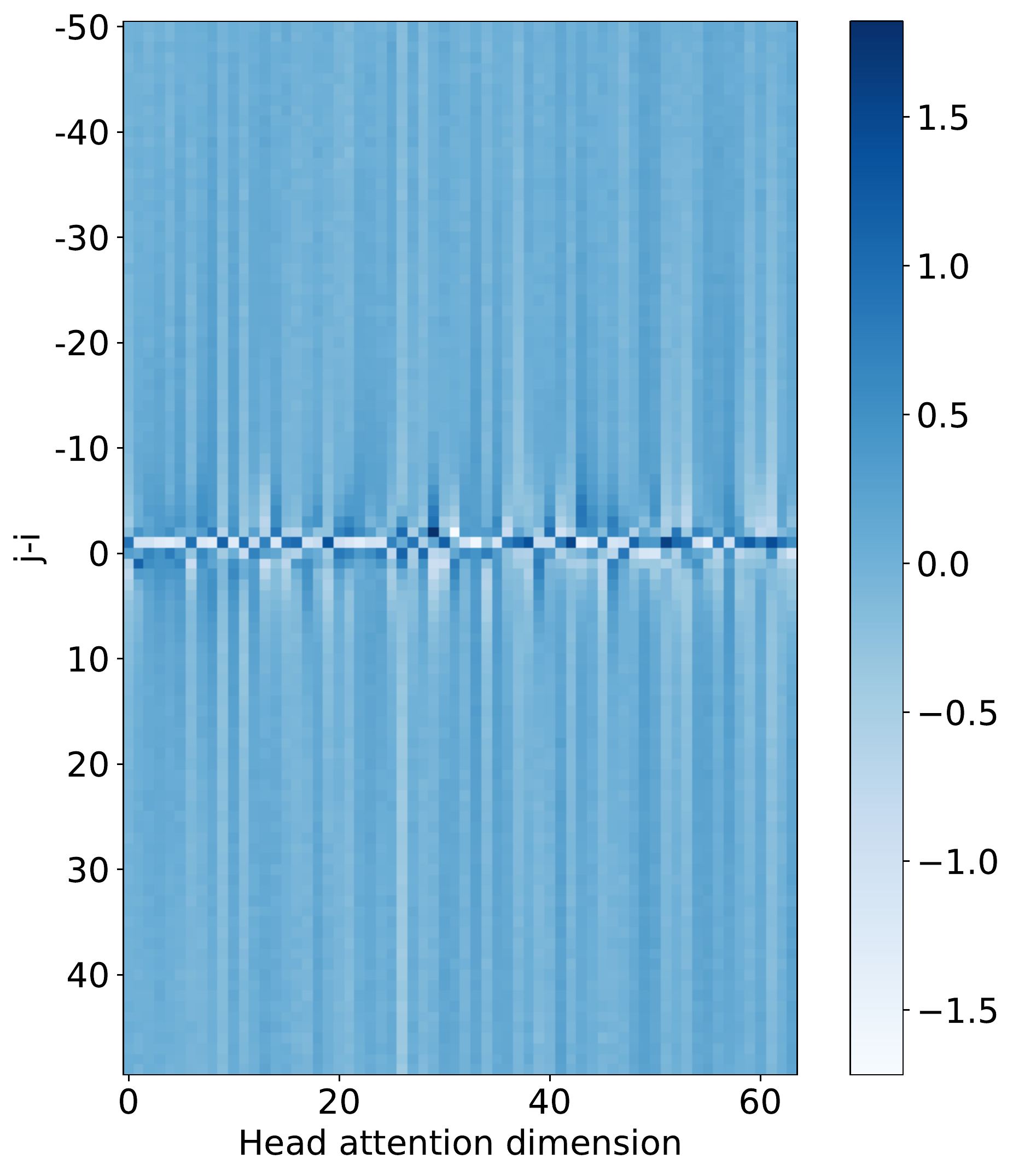}
    \caption{Relative position (from -50 to 50) embedding weights on first transformer layer and first head for method 4.}
    \label{fig:plot_emb_weights}
\end{figure}

Figure \ref{fig:plot_emb_weights} shows the embedding weights of the first head in the first layer for method 4, which is a $1023\times 64$ matrix, with the first dimension being the relative distance between two tokens, and the second being the attention dimension. We choose to plot the relative position of $[-50,50]$ which concentrates the proximity of two positions. We note that the weights at relative position of zero have the large absolute values, either positive (white) or negative (dark blue). These large absolute values may lead to large $e_{ij}$ values in equation (\ref{eqn:e_4}), which indicates a token is likely to attend to another token within a close distance.

\begin{figure}[!hbt]
    \centering
    \includegraphics[scale=0.3]{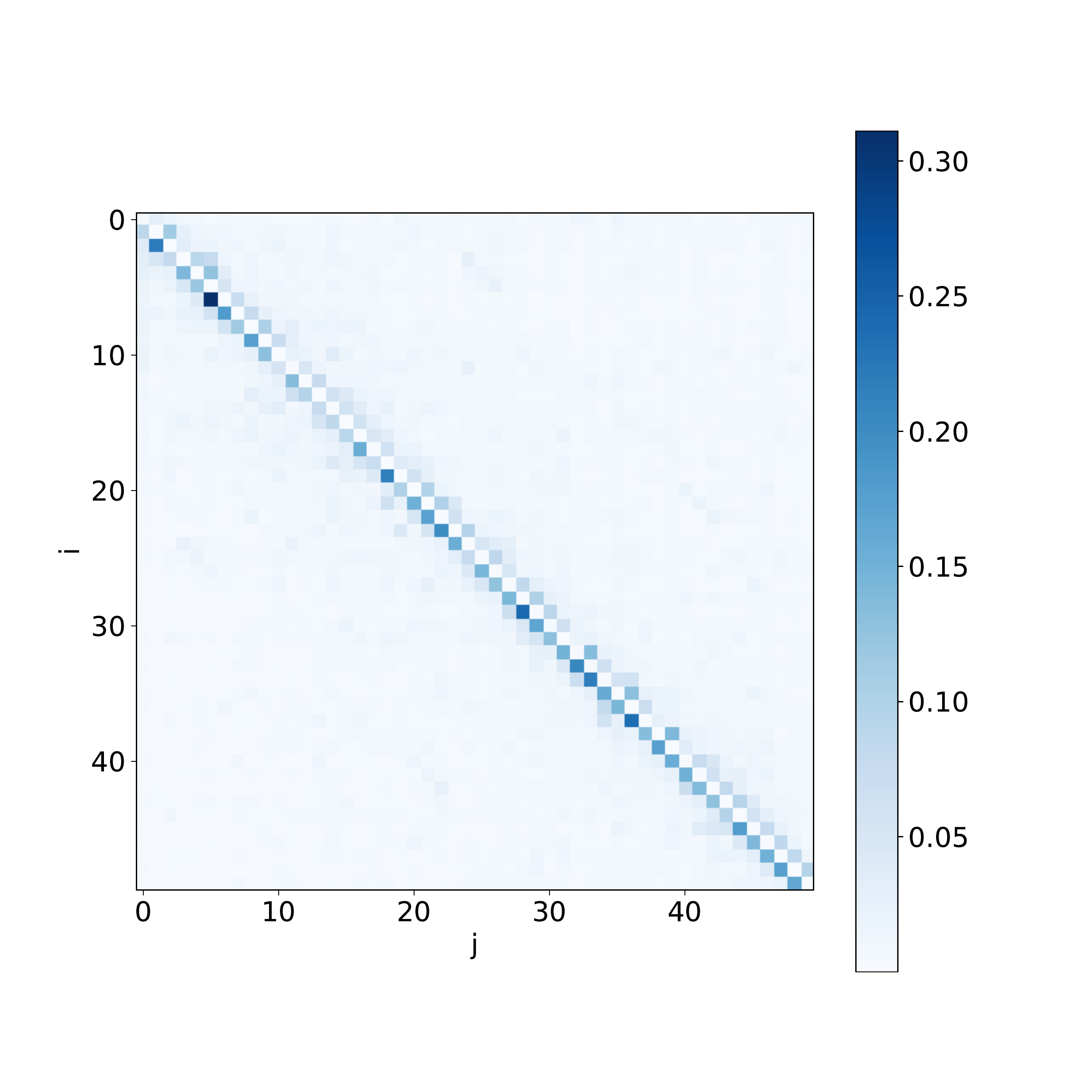}
    \caption{Averaged attention weights across 12 heads on the first transformer layer for method 4.}
    \label{fig:plot_attention}
\end{figure}

Figure \ref{fig:plot_attention} shows the averaged attention weights over 12 heads on the first transformer layer for method 4. We show the self-attention between the first 50 tokens. This clearly shows that tokens heavily attend to their neighbors (dark blue on the diagonal) and has nearly zero attentions to tokens which are far away. This also explains why a small value of $k$ is sufficient for the relative position embedding as the attention weights beyond this range are close to zero. Note that tokens usually have near zero attention on themselves. This seems counter-intuitive but can be explained by the masked language model (MLM) task, in which the neighbors of a given token (as opposed to the token itself) provide the most useful information for the task. 

\section{Conclusion}
We proposed new relative position embedding methods to encourage more interactions between query, key and relative position embeddings in self-attention mechanism. Our best proposed method is a generalization of the absolute position embedding and it leads to higher accuracy than the absolute and previous relative position embeddings on SQuAD1.1. In addition, we demonstrated empirically that our relative embedding method is reasonably generalized and robust from the inductive perspective. Finally, we showed that our proposed method can be effectively and efficiently adopted as a drop-in replacement to boost the performance of large models with a small computational budget.

\bibliographystyle{acl_natbib}
\bibliography{emnlp2020}
\appendix

\end{document}